\documentclass[10pt,twocolumn]{article}
\usepackage[letterpaper,margin=0.75in]{geometry}
\usepackage{graphicx}
\usepackage{amsmath}
\usepackage{booktabs}
\usepackage{array}
\usepackage{microtype}
\usepackage{caption}
\usepackage{newunicodechar}
\usepackage[hidelinks]{hyperref}
\usepackage{stfloats}
\newunicodechar{α}{\ensuremath{\alpha}}
\newunicodechar{β}{\ensuremath{\beta}}
\newunicodechar{μ}{\ensuremath{\mu}}
\newunicodechar{ρ}{\ensuremath{\rho}}
\newunicodechar{−}{\ensuremath{-}}
\newunicodechar{≈}{\ensuremath{\approx}}
\title{Correlated by Construction: Endpoint Choice and the Validation of Protein Screening}
\author{Eric LoPrete, Galen Loving, George Sabin, Pavle Jeremić  ·  November 2026}
\date{}
\begin{document}
\maketitle
\begin{abstract}
Machine learning for protein engineering depends on functional measurements collected at screening scale, but the designs those measurements rank are deployed in processes that run elsewhere. Whether a miniaturized assay remains informative under the conditions of use is usually assumed rather than demonstrated. We describe an automated production and screening platform and test the transfer of its solution-phase metal-affinity measurements to immobilized rare-earth separation. Primary $\mathrm{p}K_{\mathrm{d}}$-derived separation factors tracked those measured on covalently immobilized protein across 82 protein–metal-pair measurements from 35 proteins, with pairwise Spearman correlations of 0.878 to 0.923 (pooled Pearson $r$ = 0.836; Spearman ρ = 0.869). Separation factors at a fixed metal pair are insensitive to the ordering shared across the lanthanide series, so these coefficients report protein-specific correspondence directly; permutation of protein identity confirms it, giving a mean scrambled correlation of 0.003 and, for the 17 proteins measured on all three pairs, $r$ = 0.917 against 0.001 ($p$ ≈ 0.001). Profiles compared as profiles behave differently. An independent comparison against published LAMBS profiles for the same protein backbones (417 proteins, 5,212 paired measurements) reproduced normalized raw elution at $r$ = 0.870, but reassigning protein identity still reached 0.840. High profile-level agreement across assay formats is largely attributable to metal preferences shared across the series rather than to protein identity. Transfer can be demonstrated, but the endpoint decides whether a correlation shows it: a pairwise contrast removes the shared ordering; a normalized profile retains it, and needs a null that reassigns protein identity.
\end{abstract}
\section*{1. Introduction}
Machine learning can accelerate protein engineering by using measured sequence–function relationships to guide the design and selection of new candidates. Its utility depends on both the quantity of experimental data and the relevance of those data to the intended application. Broad functional screens provide a foundation for learning protein–partner interactions, while incomplete coverage can limit the performance of models trained on those measurements [1]. Generating large, consistent datasets is therefore an experimental requirement for machine-learning-guided design.

The case for measurement over computed proxies has a direct empirical basis. In a comparison of three design strategies applied to the same scaffold, structural-confidence metrics selected the panel with the weakest measured function, while a model supervised on measured affinity ranked designs reliably [2]. Direct supervision on measured function is the stronger selection signal, which places the burden on the measurement itself.

That burden has a second part, less often examined. Rapid industrial deployment requires that properties measured in a miniaturized assay remain informative as proteins advance toward a product. Primary assays can measure thousands of protein–partner combinations under controlled conditions, but development commonly changes protein production, presentation and operating conditions. A screening measurement that ranks designs correctly in solution is useful only if that ranking survives immobilization, competition and the other conditions of the process it is meant to inform.

Demonstrating that it does is harder than it appears. The natural test is to compare screening measurements against an application-relevant endpoint, often one published by another group, and to report the correlation between them. Where the measured entities form a chemically ordered family, this comparison carries a trap. Members of such a family share a systematic response ordering, and any two profiles drawn from the family will agree on that ordering whether or not they describe the same protein. A high correlation between predicted and observed profiles can therefore be produced almost entirely by shared chemistry, with the correlation reporting the family trend rather than the agreement between measurements on a given protein. Unless such a comparison is evaluated against a null that breaks the correspondence between each profile and its own protein, it does not establish what it is taken to establish.

Rare-earth separation provides a practical setting for both questions: whether a screening measurement stays informative at process scale, and how that is demonstrated. It is also an unusually clear example of the trap. Rare-earth-containing permanent magnets are important to electric motors, wind energy and electronics, motivating interest in reliable supplies and recovery routes [3, 4]. The chemical similarity of neighboring rare-earth elements makes selective separation challenging, and conventional refining can require extensive processing [5]. Proteins offer a sequence-programmable approach to metal recognition. Their value for separation depends on retaining useful selectivity when produced at larger scale, immobilized and exposed to competing metals. At the same time, the lanthanide contraction imposes a strong shared ordering across the series: affinity profiles of lanthanide-binding proteins tend to rise through the light elements, peak near the middle of the series and decline toward the heavy elements, whatever the protein.

Recent work on lanmodulin provides an independent connection between protein recognition and separation behavior. Diep and colleagues developed SpyTag-Catcher Immobilization of Lanmodulin for Assaying Metal-Binding Selectivity (SpyCI-LAMBS), characterizing 621 lanmodulin orthologs against 15 rare-earth elements and evaluating selected proteins in chromatographic separations [6]. These measurements offer an orthogonal test of whether solution-phase affinities capture metal preferences observed in an immobilized competitive assay.

At Aether, we have developed an automated platform linking individually tracked constructs, cell-free production, purification, tag-based quantification and concentration-normalized primary screening. Selected proteins advance to milligram-scale production, covalent immobilization and competitive metal-separation assays. Here we compare primary $\mathrm{p}K_{\mathrm{d}}$-derived selectivity against separation factors measured on immobilized protein, and project the same primary affinities onto published LAMBS raw-elution and LogD profiles. For both comparisons we evaluate the observed correspondence against a permutation null in which protein identity is reassigned, isolating the component of agreement that depends on matching each protein's screening measurement to its own scale-up result. The two comparisons give different answers, and the difference turns on how each endpoint is constructed.
\section*{2. High-throughput protein data production and screening platform}
Key to building our machine learning models is the regular supply of high-quality experimental data generated across hundreds of proteins each week. To support this, we have built an end-to-end, high-throughput protein production and screening platform designed to be deployed across multiple internal protein engineering programs. The broader goal is to continually expand both the scale and diversity of experimental data available to our models and, ultimately, the range of relationships between protein sequence and function that they can learn. The metal-binding work described here represents one example of how this general capability can be applied to a specific protein engineering problem. In this application, the platform allows us to routinely collect experimentally determined binding constants across thousands of protein–metal pairings every week.

\subsection*{2.1 Protein design and production}
The process begins at the computer, where we use the latest version of our Protein Function Model (PFM) to design hundreds of novel protein constructs for testing. Our LIMS tools are then used to generate a set of codon-optimized DNA sequences, which are ordered from a commercial vendor as individually synthesized constructs arrayed in plates, with each well containing a single construct.

Upon arrival, the DNA constructs are amplified using established molecular biology techniques and transferred into deep-well plates containing cell-free extract prepared in house from cell lysates under conditions that preserve the cellular machinery necessary for protein synthesis (e.g., ribosomes, tRNAs, and the enzymes responsible for charging tRNAs with their corresponding amino acids). Each construct is maintained separately throughout protein production and screening to enable individual evaluation, without pooling samples at any stage. Following expression, the proteins are isolated using affinity purification and desalting techniques carried out as high-throughput, plate-based workflows on our automated platform. Isolated yields generally range from tens to hundreds of micrograms per construct, depending on the expression level of the individual protein.

\subsection*{2.2 Protein quantification and assay preparation}
At this point, the library consists of purified protein stocks at varying concentrations arrayed across multiple plates. The next step is protein quantification. Because our goal was to build a screening platform that could accommodate as broad a range of proteins as possible, we needed a quantification method that did not depend strongly on the sequence or structural properties of the protein itself. We therefore developed a method based on a conserved, protease-cleavable tag present on all constructs. Following proteolysis, the resulting product is quantified using an Agilent RapidFire system coupled to a triple-quadrupole mass spectrometer. Proteolysis, sample preparation, and mass-spectrometric analysis of an entire production unit, the set of hundreds of novel protein constructs processed together in one work-cell, can be completed in less than eight hours.

With accurate concentration measurements for each construct, we can normalize protein concentrations across the unit and generate multiple copies of each plate, providing a collection of assay-ready plates for downstream screening. At our current scale, the entire process from receipt of synthesized DNA to assay-ready protein plates can be completed in approximately two business days. For example, DNA received on Monday morning can be ready for protein screening by Wednesday morning.

\subsection*{2.3 Primary screening}
Primary screening is performed in a highly parallelized fashion using automated work cells equipped for plate-based workflows. These include nanoliter-scale liquid handlers paired with multiple plate readers, allowing different parts of the workflow to operate in parallel while minimizing downtime of non-bottleneck systems.

Our primary binding assays take advantage of changes in the natural photophysical properties of certain metal ions that occur upon interaction with a protein binding partner. By measuring how these signals change as a function of the concentration of one of the binding partners, the resulting data can be fit to appropriate binding models to determine dissociation constants. These measurements allow us to characterize binding across thousands of protein–metal combinations and infer separation factors between closely related metals.

\subsection*{2.4 Secondary screening}
Proteins that emerge from primary screening with interesting properties are nominated for more detailed secondary evaluation. These include proteins showing promising metal-binding selectivity as well as proteins that provide useful benchmarks for understanding structure–activity relationships. For secondary screening, selected proteins are modified to introduce structural elements that enable site-specific covalent immobilization through a discrete amino acid residue, while the quantification tags required for primary screening are removed. These proteins are expressed in transformed BL21 cells at milligram scale and purified using standard affinity purification methods.

Purified proteins are then covalently immobilized on an agarose-based resin containing reactive groups that selectively couple to complementary groups on the protein. Secondary screening directly measures separation factors for selected pairs of metal ions using competition experiments. Two (or more) metals are combined at equal molar ratios and in large molar excess over the available binding sites on the immobilized protein. The metal mixture is incubated with the protein resin under binding conditions, generally at pH 5 or above, for sufficient time to reach equilibrium. The resin is then washed with binding buffer to remove unbound metal ions, followed by elution under acidic conditions at pH below 2.

All fractions are collected and the concentrations of the individual metals are quantified by ICP-OES. These measurements allow us to directly determine effective separation factors under competitive binding conditions. The separation factors measured in these direct challenge experiments can then be compared with those inferred from the independently measured dissociation constants generated during primary screening. This provides an experimental link between the high-throughput binding measurements used to generate large datasets and the separation behavior of selected proteins under conditions that more closely approximate their intended application.

Together, the primary and secondary screening workflows provide complementary layers of information. Primary screening provides the throughput needed to explore large numbers of protein–metal interactions and generate the datasets used to improve our models, while secondary screening provides a direct experimental test of whether the binding behavior identified in primary screening translates into useful metal-separation performance. The resulting experimental data are then returned to the computational workflow to inform subsequent rounds of protein design.

\subsection*{2.5 Signal processing}
Negative and positive controls spanning three orders of magnitude in affinity are included across all measured metals to account for process noise, monitor screening consistency, detect activity and normalize measured activity. Affinities are expressed as $\mathrm{p}K_{\mathrm{d}} = -\log_{10}(K_{\mathrm{d}}/1\,\mathrm{M})$. To reduce pointwise variation, primary $\mathrm{p}K_{\mathrm{d}}$ profiles are smoothed with a quadratic function of ionic radius [7]. The smoothed values are used to calculate primary separation factors, $\alpha_{A/B} = 10^{\,\mathrm{p}K_{\mathrm{d}}^{A} - \mathrm{p}K_{\mathrm{d}}^{B}}$, and to compare with secondary screening and LAMBS measurements.

\subsection*{2.6 Comparison with LAMBS}
Expression concentrations were summarized across the full production panel, counting each distinct Aether construct once irrespective of its production unit or binding-label status. For the affinity–elution comparison, only accepted numeric $\mathrm{p}K_{\mathrm{d}}$ measurements and profiles with sufficient metal coverage for smoothing and comparison were retained. Concentration-related label quality checks and unsuccessful or insufficient affinity determinations reduced the evaluable cohort. The 20 μM line summarizes the expression distribution and does not impose an additional cutoff on the correlation analysis.

Published LAMBS data provide an orthogonal scale-up comparison for the same protein backbones, with SpyTag-mediated immobilization in the LAMBS assay [6]. Smoothed primary $\mathrm{p}K_{\mathrm{d}}$ values are converted to $K_{\mathrm{d}}$ and used in an equilibrium approximation to predict fractional metal retention, $f_i = L/(K_{\mathrm{d},i} + L)$, and $\log_{10} D_i = \log_{10}(1 + L/K_{\mathrm{d},i})$, where $L$ is an effective concentration of available binding sites. Predicted retention is compared with mean raw elution, and predicted LogD with published LogD. Each profile is normalized to its maximum over the metals included in the comparison. The effective binding-site concentration is calibrated separately for each endpoint using other proteins, excluding all reference profiles corresponding to the protein being evaluated. Spearman correlations summarize pairwise and overall correspondence; Pearson correlations use log-transformed separation factors for Aether and normalized elution or LogD for LAMBS.

The published total recovered metal, $M_{\mathrm{tot}}$, was used as a joint expression-and-binding proxy to assess LAMBS data quality [6]. A background QC subset required all three replicate totals to exceed the reported nonspecific background of 6.4 ppb. Both projections were recalibrated within this subset, excluding the evaluated protein, and compared alongside the full cohort.

\subsection*{2.7 Permutation tests of protein assignment}
Correlations between screening-derived and scale-up measurements can arise from two sources. The first is protein-specific: the screening measurement for a given protein matches the scale-up measurement for that same protein. The second is shared across the series: all lanthanide-binding proteins show a broadly similar dependence of affinity on ionic radius, so two profiles drawn from different proteins will still agree on that dependence. Only the first source supports the use of screening measurements to rank candidates.

Whether the second source can contribute at all depends on how the endpoint is constructed. Write a primary measurement as a shared series trend plus a protein-specific deviation, $\mathrm{p}K_{\mathrm{d},p}^{A} = g^{A} + u_{p}^{A}$. The log separation factor for a fixed metal pair is then $\log_{10}\alpha_{p,A/B} = (g^{A} - g^{B}) + (u_{p}^{A} - u_{p}^{B})$. The shared term is the same for every protein, and a correlation computed across proteins is invariant to an additive constant. A separation factor at a fixed metal pair therefore carries no contribution from the shared ordering, by construction. A profile compared as a profile retains that ordering in full, and normalization to a profile maximum does not remove it.

The two sources are not distinguished by the correlation coefficient itself. Where an endpoint admits the second source, permutation of protein identity measures how much of the agreement depends on correct matching. Where it does not, the same permutation serves as a check that pooling across metal pairs has introduced no shared structure, since the pairs differ in their offsets.

We recomputed each comparison under 1,000 random reassignments of protein identity. In each reassignment, whole primary profiles were exchanged among proteins, preserving metal identities and the number of measurements contributing to each comparison. Shuffling was restricted to groups of proteins with identical measurement coverage and identical numbers of associated reference profiles, so that no reassignment altered the composition of the pooled dataset. Smoothing, normalization and the calibrated projections were held fixed at their values for the correct assignment, so that the reassignment changed only which scale-up measurement each primary profile was compared against. Where a protein's coverage pattern was unique within its cohort, no exchange partner existed and that protein remained fixed; this applied to 1 protein in the Aether secondary-screening comparison and 43 proteins in the full LAMBS cohort. Multiple LAMBS reference profiles representing the same Aether construct were kept together.

Because fixed proteins contribute their correct assignment to every permutation, they bias the scrambled distribution toward the observed value. We therefore repeated each test on the subset of proteins with complete measurement coverage, where every protein can exchange assignments with every other: 17 proteins measured on all three metal pairs for the Aether comparison, and 93 proteins measured on all 15 metals for the LAMBS comparison.

One-sided permutation values are reported as $p = (b + 1)/(1{,}000 + 1)$, where $b$ counts scrambled correlations at least as large as the observed value, without adjustment for multiple comparisons. Permutation tests were applied to pooled Pearson and Spearman correlations and to the individual pairwise rank correlations.
\section*{3. Results}
\subsection*{3.1 Primary affinities correspond to secondary separation factors, and the correspondence depends on protein identity}
The paired dataset contained 82 protein–metal-pair measurements from 35 proteins with accepted primary affinities and secondary-screening separation factors. Primary $\mathrm{p}K_{\mathrm{d}}$-derived separation factors tracked those measured in secondary screening across Dy/Pr, Pr/Nd and Tb/Dy (Figure 1). All three metal pairs showed strong rank correspondence, with Spearman correlations from 0.878 to 0.923 (Table 1). Across the pooled measurements, Spearman ρ was 0.869 and Pearson $r$ on log-transformed factors was 0.836 (Table 2).

\begin{figure}[tbp]
\centering
\includegraphics[width=\columnwidth]{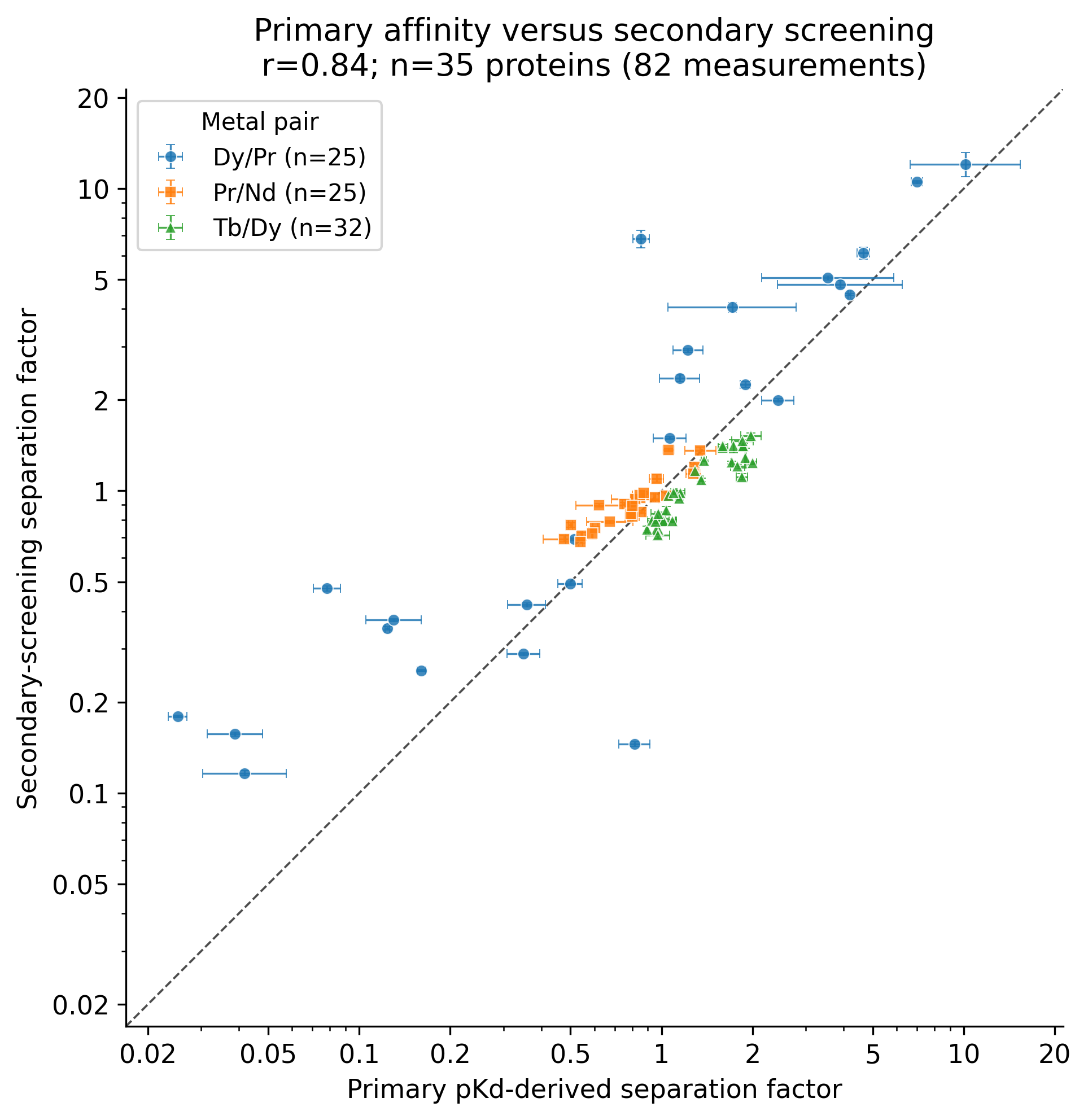}
\caption{Figure 1. Primary affinity-derived selectivity corresponds to immobilized secondary-screening selectivity. Each point represents one protein–metal-pair measurement; colors identify Dy/Pr, Pr/Nd and Tb/Dy. The horizontal axis shows the separation factor calculated from smoothed primary $\mathrm{p}K_{\mathrm{d}}$ values, and the vertical axis shows the secondary factor calculated from collected fractions. Both axes are logarithmic. The title reports Pearson $r$ on log-transformed factors and the number of proteins, $n$. Horizontal bars propagate reported primary affinity spread through signal processing; vertical bars show secondary analytical error.}
\label{fig:secondary_scatter}
\end{figure}

\begin{table*}[tbp]
\centering\small
\caption{Table 1. Spearman correlations across proteins for matched metal pairs. Parentheses give the number of paired comparisons: protein measurements for Aether and reference-profile pairs for LAMBS. Aether compares primary and secondary separation factors. LAMBS compares predicted and observed A/B ratios of each profile endpoint, which are distinct from separation factors. Background QC requires all three total-metal replicates above 6.4 ppb. Pairwise coefficients test ranking for a fixed metal pair; Table 2 pools measurements across metals.}
\label{tab:pair_correlations}
\setlength{\tabcolsep}{4pt}
\begin{tabular}{@{}lccc@{}}
\toprule
Comparison & Dy/Pr & Pr/Nd & Tb/Dy \\
\midrule
Aether secondary screening & 0.878 (25) & 0.923 (25) & 0.896 (32) \\
LAMBS raw elution & 0.331 (405) & 0.610 (354) & −0.056 (415) \\
LAMBS LogD & 0.422 (405) & 0.641 (354) & −0.045 (415) \\
LAMBS raw elution (background QC) & 0.371 (348) & 0.700 (302) & −0.121 (357) \\
LAMBS LogD (background QC) & 0.462 (348) & 0.702 (302) & −0.127 (357) \\
\bottomrule
\end{tabular}
\end{table*}

The agreement spans changes from cell-free production and soluble binding measurements to milligram-scale production and covalent immobilization. Individual deviations and the reported uncertainty remain visible in the pooled scatter, while the consistent ordering across the three pairs supports using primary affinity measurements to prioritize proteins for secondary evaluation.

The 35 proteins evaluated by Aether secondary screening showed diverse primary-affinity profiles across the REE series (Figure 2). Their normalized retention projections, based on 434 accepted metal-affinity measurements, varied broadly in profile shape. This diversity contrasts with the predominant light-REE preference in the LAMBS cohort (Figure 4).

\begin{figure}[tbp]
\centering
\includegraphics[width=\columnwidth]{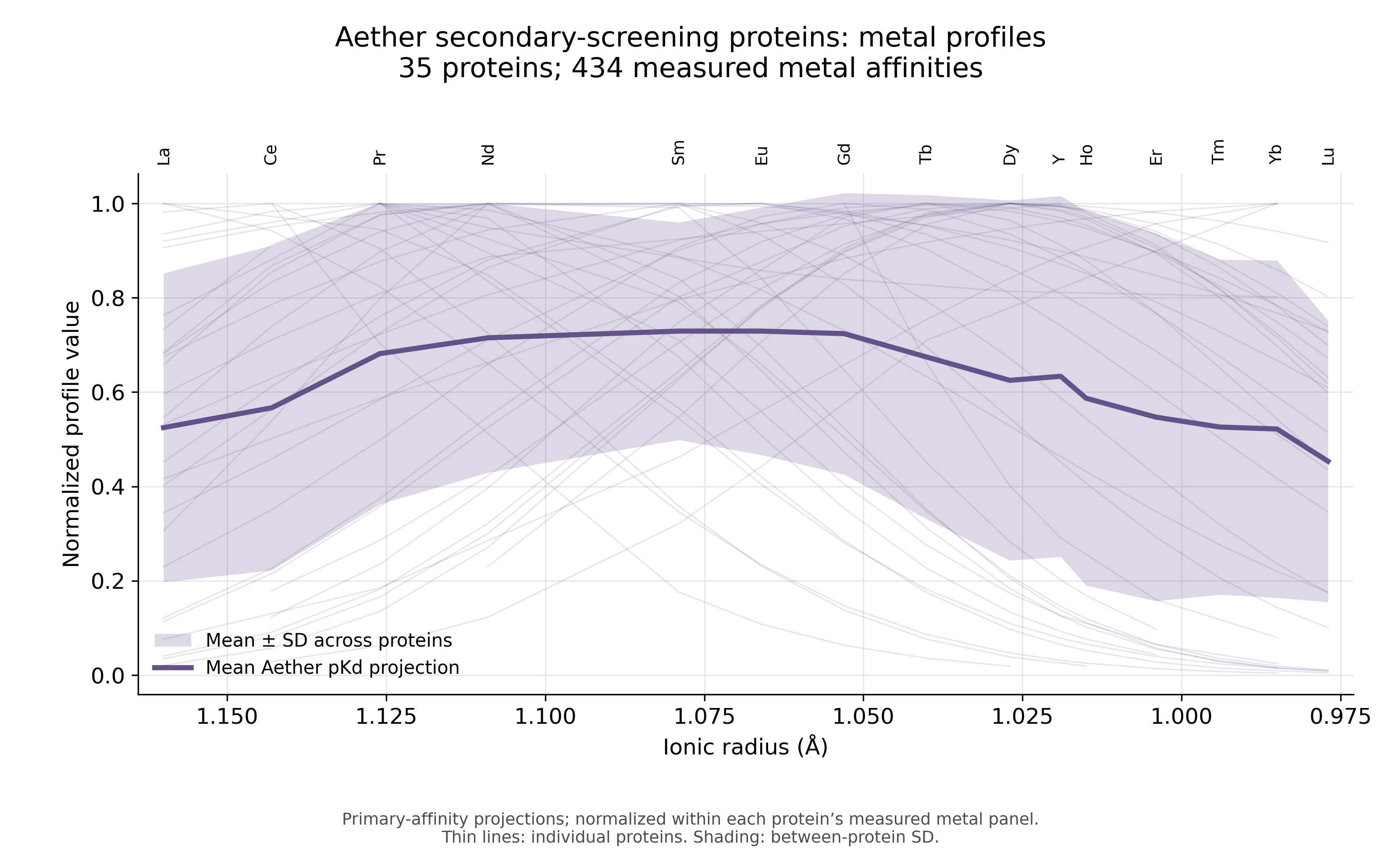}
\caption{Figure 2. Diverse metal-affinity profiles among the 35 proteins evaluated in Aether secondary screening. Thin purple lines show individual profiles projected from smoothed primary $\mathrm{p}K_{\mathrm{d}}$ values using the fractional-retention transformation; the thick line and shaded band show the mean and one standard deviation across proteins. The display uses a common effective binding-site concentration fixed to the median calibration value from the LAMBS raw-elution comparison. Each protein's profile is normalized to its maximum over its measured metal panel, and each metal summary includes proteins with accepted measurements for that metal. The profiles contain 434 measured protein–metal affinities.}
\label{fig:aether_metal_profiles}
\end{figure}

Separation factors at a fixed metal pair are insensitive to the shared series ordering by construction (Section 2.7), so the pairwise coefficients above cannot have been inflated by it. Pooling across three metal pairs is a different matter: the pairs differ in their shared offsets, and those differences could survive reassignment. We therefore evaluated the comparison against the permutation null described in Section 2.7. Under 1,000 reassignments of protein identity, the pooled Pearson correlation fell from $r = 0.836$ to a scrambled mean of 0.003 (central 95\% permutation interval, −0.320 to 0.381; Figure 3). Pooled Spearman correlation fell from ρ = 0.869 to a scrambled mean of 0.097. Both pooled correlations and the individual Dy/Pr, Pr/Nd and Tb/Dy rank correlations exceeded all 1,000 scrambled values (one-sided permutation $p$ ≈ 0.001). In the complete-panel test, where every protein can exchange assignments, the 17 proteins measured on all three pairs gave a pooled Pearson correlation of 0.917 against a scrambled mean of 0.001 ($p$ ≈ 0.001).

The pooled scrambled mean of 0.003 shows that the three pairs introduce no shared structure of their own, so the pooled coefficient reports the same protein-specific correspondence as the pairwise ones. Agreement between the primary screen and the immobilized separation assay rests on correctly identifying the protein associated with each selectivity profile.

\begin{figure*}[tbp]
\centering
\includegraphics[width=1.0\textwidth]{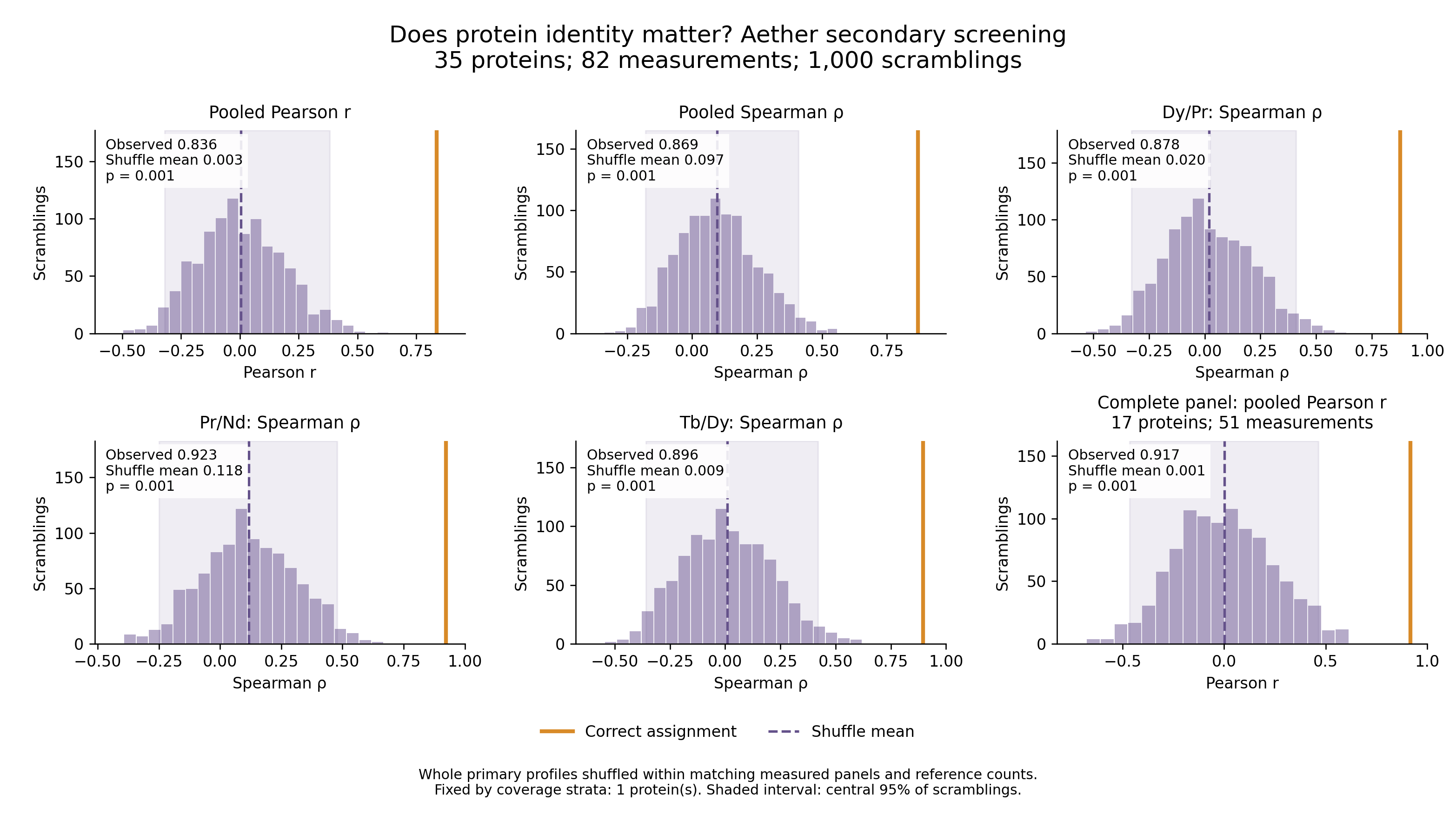}
\caption{Figure 3. Correspondence with Aether secondary screening depends on protein assignment. Histograms show correlations from 1,000 whole-profile reassignments; orange lines mark correctly matched correlations and dashed purple lines mark scrambled means. Shaded intervals contain the central 95\% of scrambled correlations. Panels show pooled Pearson correlation on log-transformed separation factors, pooled and pairwise Spearman correlations, and pooled Pearson correlation for the 17 proteins measured for all three pairs. One protein with unique coverage remains fixed in the full-cohort test; every protein can exchange assignments in the complete-panel test. One-sided permutation values are calculated as $p = (b + 1)/(1{,}000 + 1)$, where $b$ counts scrambled correlations at least as large as the observed value, without adjustment for multiple comparisons.}
\label{fig:aether_permutation}
\end{figure*}

\subsection*{3.2 Profile-level agreement with external LAMBS data is dominated by shared series chemistry}
Of the 610 constructs in the LAMBS production panel, 417 provided accepted affinities and sufficient metal coverage for comparison with 419 published reference profiles across 5,212 paired metal measurements. The affinity-based retention projection reproduced normalized LAMBS raw-elution profiles with Pearson $r = 0.870$ and Spearman ρ = 0.859 (Figure 4; Table 2). Projection onto the published normalized LogD endpoint gave $r = 0.858$ and ρ = 0.849. Background QC retained 357 proteins and 4,448 paired metal measurements and raised both endpoints further, to $r = 0.903$ and $r = 0.895$ respectively.

Taken at face value, these are high correlations between independently generated measurements on the same protein backbones. The permutation test shows that they are not what they appear to be.

\begin{figure*}[tbp]
\centering
\includegraphics[width=1.0\textwidth]{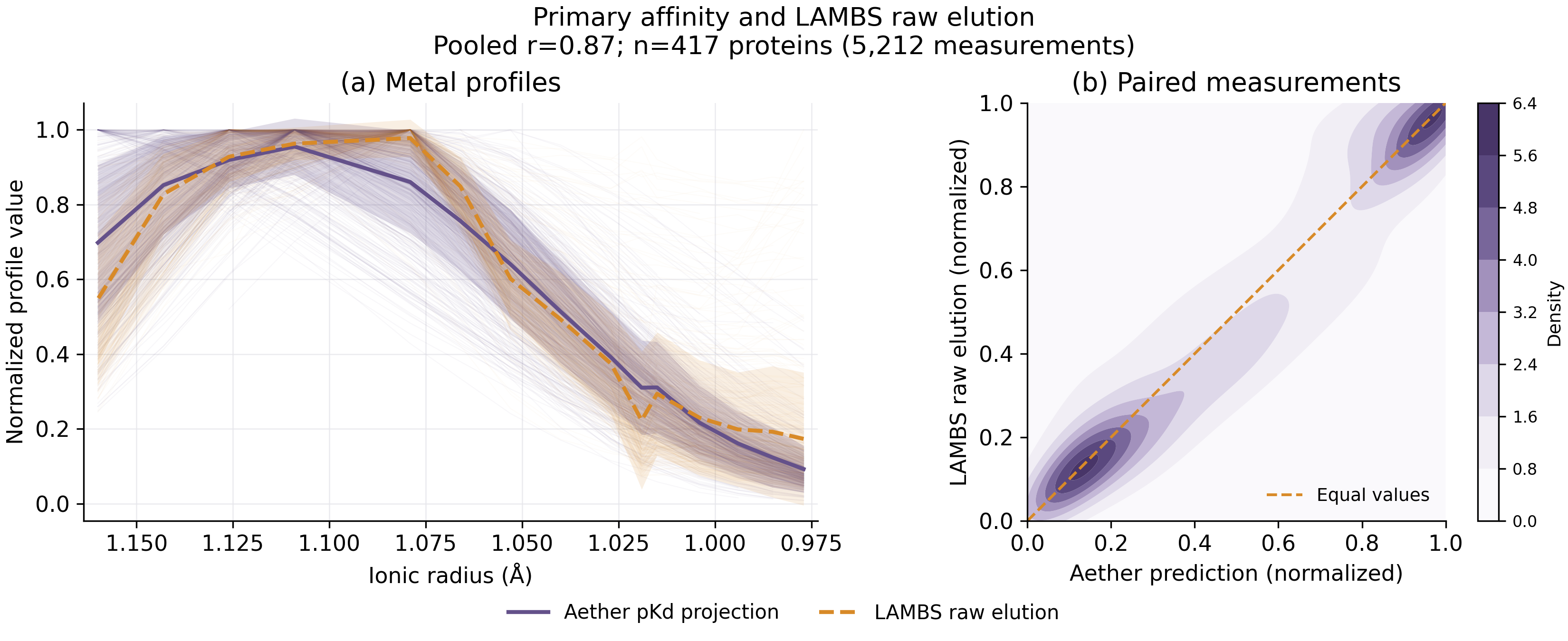}
\caption{Figure 4. Primary-affinity projections and LAMBS raw elution for the full affinity-comparison cohort. (a) Profiles across ionic radius: faint traces show individual reference profiles; solid purple and dashed orange curves show mean Aether retention predictions and LAMBS raw elution, respectively, with shaded bands showing one standard deviation across profiles. At each metal, both summaries use the available paired measurements. (b) The same paired measurements plotted with Aether prediction on the horizontal axis and LAMBS raw elution on the vertical axis. Purple shading shows their smoothed density, with darker regions indicating greater concentration of measurements; the dashed orange line indicates equal values. Each profile is normalized to its maximum over the metals included in the comparison, with all profiles corresponding to the evaluated protein excluded from calibration. The title reports Pearson $r$ across all 5,212 paired measurements and the number of distinct Aether proteins, $n$.}
\label{fig:lambs_overview}
\end{figure*}

\begin{table*}[tbp]
\centering\small
\caption{Table 2. Overall correspondence between primary affinities and immobilized-assay endpoints, with the corresponding permutation nulls. Pearson $r$ uses log-transformed separation factors for Aether and normalized raw-elution or LogD values for LAMBS. Both LAMBS endpoints use the same paired measurements within each cohort. Background QC requires all three total-metal replicates above 6.4 ppb; calibration excludes all profiles of the evaluated protein within each cohort. Counts are distinct Aether proteins and paired endpoint measurements, retaining corresponding LAMBS reference profiles when more than one represents the same Aether construct. Mean scrambled correlations summarize 1,000 whole-profile reassignments within matching measurement-coverage and reference-count groups, with the corresponding endpoint- and cohort-specific projections held fixed (Section 2.7). The difference between the correct-assignment and mean-scrambled columns is the component of agreement attributable to protein identity. Figures 3 and 5 show the Aether and full-cohort LAMBS raw-elution tests, respectively.}
\label{tab:overall_correlations}
\setlength{\tabcolsep}{4pt}
\begin{tabular}{@{}lrrrrrr@{}}
\toprule
Comparison & Proteins & Measurements & Correct $r$ & Correct ρ & Scrambled $r$ & Scrambled ρ \\
\midrule
Aether secondary screening & 35 & 82 & 0.836 & 0.869 & 0.003 & 0.097 \\
LAMBS raw elution & 417 & 5,212 & 0.870 & 0.859 & 0.840 & 0.829 \\
LAMBS LogD & 417 & 5,212 & 0.858 & 0.849 & 0.826 & 0.817 \\
LAMBS raw elution (bg QC) & 357 & 4,448 & 0.903 & 0.889 & 0.875 & 0.861 \\
LAMBS LogD (bg QC) & 357 & 4,448 & 0.895 & 0.884 & 0.865 & 0.855 \\
\bottomrule
\end{tabular}
\end{table*}

Under 1,000 reassignments of protein identity, the mean scrambled Pearson correlation was $r = 0.840$ (central 95\% permutation interval, 0.834 to 0.847), against 0.870 for correct assignments (Figure 5). Mean scrambled Spearman correlation was ρ = 0.829, against 0.859. The same pattern held for the LogD endpoint and for the background-QC cohort, where the scrambled means were 0.875 and 0.865 against observed values of 0.903 and 0.895 (Table 2). Correct matching exceeded all 1,000 scrambled values in every case ($p$ ≈ 0.001), so the protein-specific component is detectable, but it accounts for roughly 0.03 in correlation units out of an observed 0.87. Most of the agreement is reproduced by comparing each primary profile against a different protein's elution data.

The complete-panel test gave the same answer with every protein free to exchange assignments: for the 93 LAMBS proteins measured on all 15 metals, the observed pooled Pearson correlation was 0.872 against a scrambled mean of 0.839 ($p$ ≈ 0.002). The result is therefore not an artifact of the 43 proteins held fixed by unique coverage patterns in the full-cohort test.

\begin{figure*}[tbp]
\centering
\includegraphics[width=1.0\textwidth]{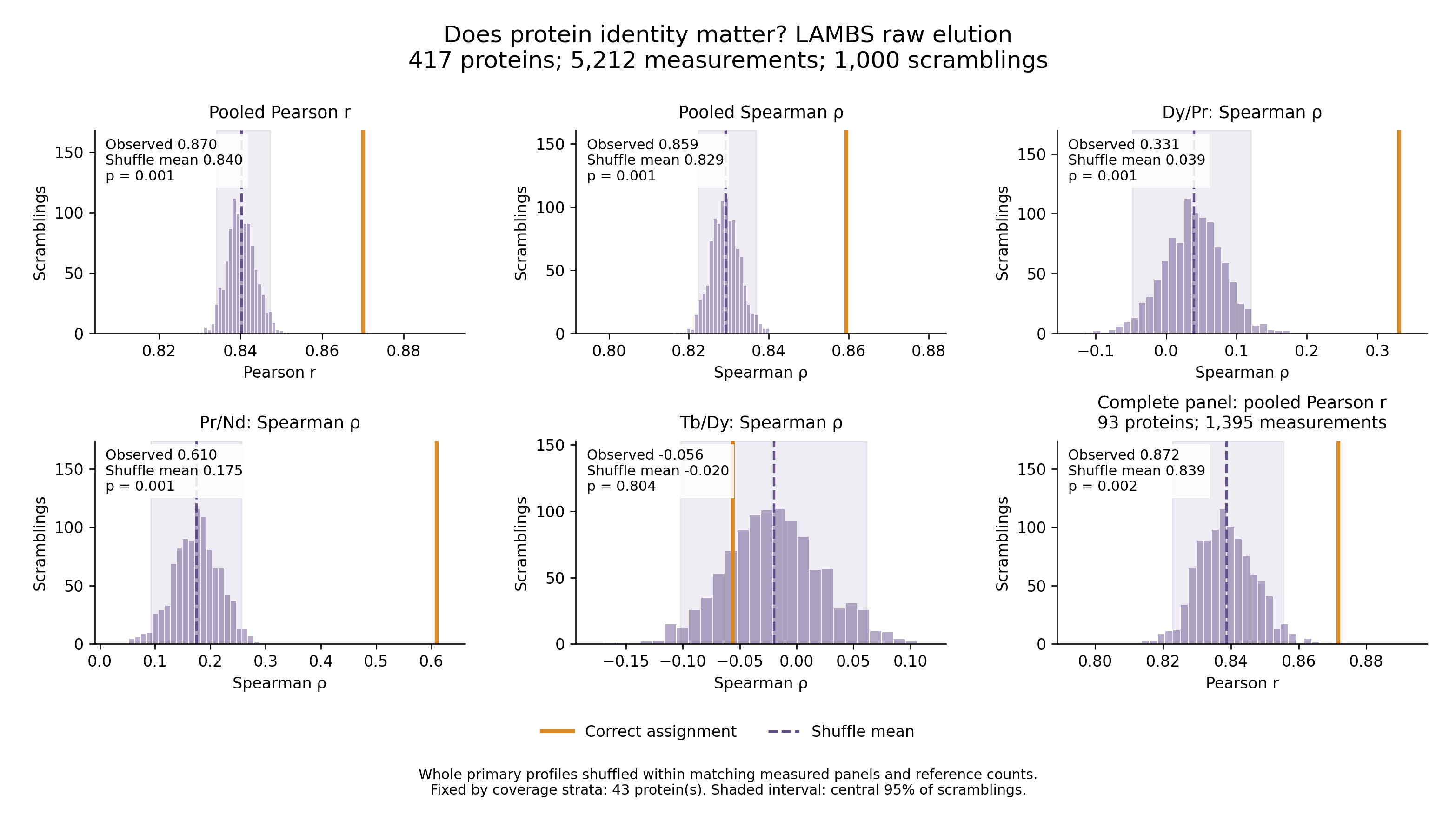}
\caption{Figure 5. LAMBS profiles retain high pooled correlation under scrambled protein assignments. Plot conventions and permutation $p$ values follow Figure 3. Pooled correlations compare normalized primary retention projections with normalized LAMBS raw elution; pairwise correlations compare their metal A/B ratios. The full-cohort test retains 417 proteins and 5,212 measurements, with 43 proteins fixed by unique combinations of metal coverage and reference count. Multiple reference profiles for the same Aether construct remain together. The complete-panel test includes 93 proteins measured for all 15 metals, allowing every protein to exchange assignments. Correct matching improves pooled agreement and Dy/Pr and Pr/Nd ranking, while Tb/Dy remains within the scrambled distribution.}
\label{fig:lambs_permutation}
\end{figure*}

The pairwise tests separate the comparison further. Correctly matched Dy/Pr and Pr/Nd rank correlations exceeded all 1,000 scrambled values, so protein identity carries information for those pairs even though the pairwise coefficients themselves are modest (0.331 and 0.610; Table 1). Tb/Dy showed no matching advantage ($p = 0.804$): the observed rank correlation of −0.056 lies within the scrambled distribution, and background QC did not change this (−0.121). We therefore do not interpret the Tb/Dy comparison against LAMBS in either direction. The same pair is resolved protein-specifically in Aether secondary screening, where its rank correlation of 0.896 exceeded all scrambled values.

This pattern is the reason the permutation null is necessary rather than optional. The pooled LAMBS correlation of 0.87 and the Tb/Dy rank correlation of −0.06 are computed from the same measurements on the same proteins. Without a null, the first invites the conclusion that the primary assay transfers to the LAMBS format and the second invites a search for a mechanism explaining why one metal pair fails. Against the null, both resolve: the pooled figure mostly reports the lanthanide contraction, and the pairwise figure reports an absence of protein-specific signal that requires no further explanation.

\subsection*{3.3 Protein production spans diverse structural classes}
A separate expression panel comprised 78 non-LanM proteins, of which approximately three quarters reached concentrations of at least 150 μg/mL. Structural identifiers were available for 74 of the 78, allowing their placement in a fold-space map of 15,008 nonredundant CATH S20 domains [8, 9] (Figure 6). The tested proteins span mainly α-helical, mainly β-sheet and mixed α/β regions, demonstrating expression across a broad range of soluble protein structures. This panel is reported in mass concentration because it was quantified independently of the LAMBS production panel described below, which is reported in molar concentration.

\begin{figure*}[tbp]
\centering
\includegraphics[width=1.0\textwidth]{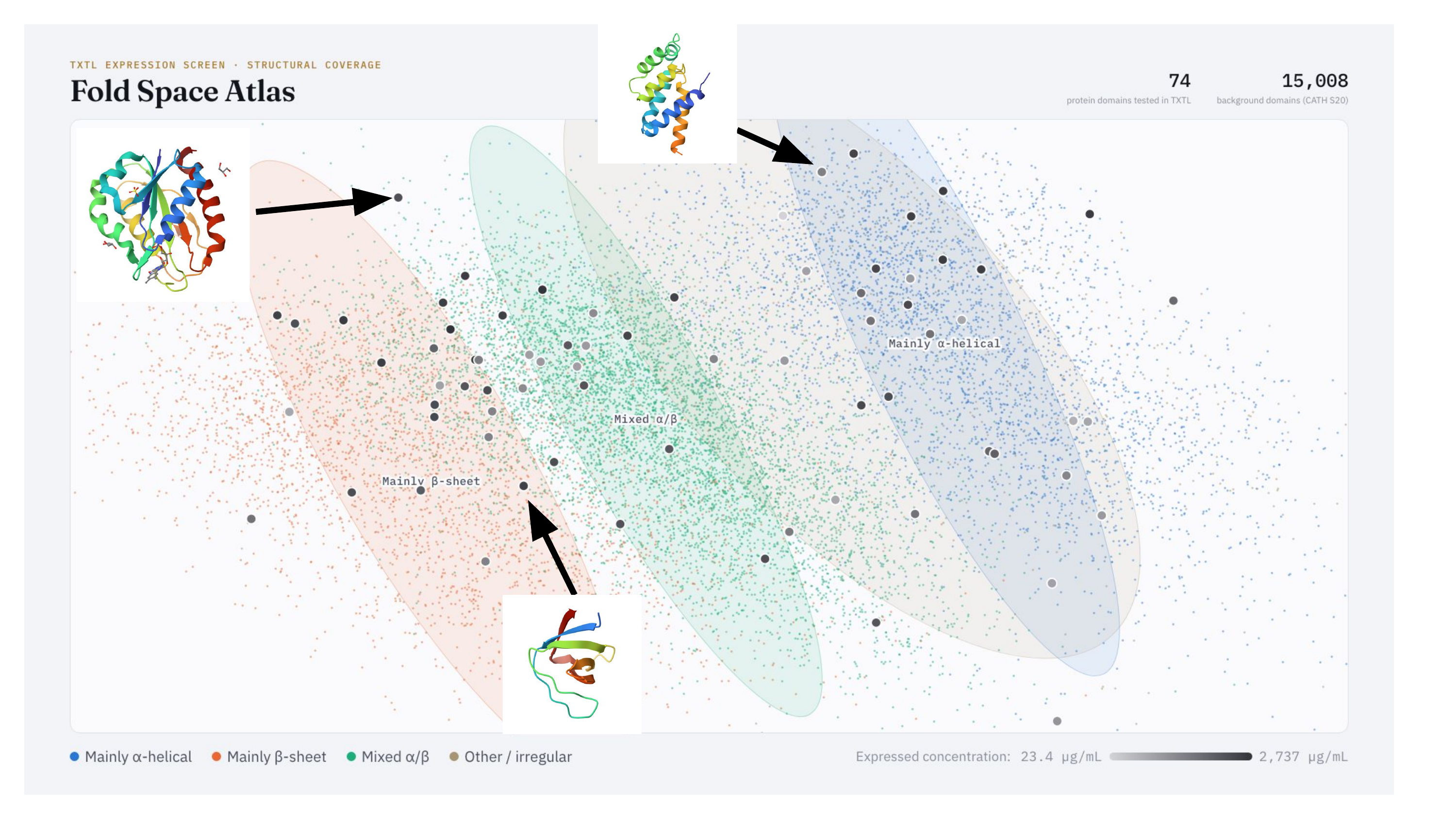}
\caption{Figure 6. Fold-space coverage of proteins evaluated in cell-free expression. Small background points represent 15,008 nonredundant CATH S20 domains, colored by structural class: mainly α-helical (blue), mainly β-sheet (orange), mixed α/β (green), and other or irregular (brown). Axes are the first two principal components of background-domain log sequence length, isoelectric point, net charge, hydrophobicity, and helix and sheet content. The 74 tested protein domains are projected into the same space and shown as large circles; darker grayscale shading indicates higher measured expression concentration, spanning 23.4–2,737 μg/mL. The colored ellipses summarize the background classes, and arrows link representative structures to their positions. The figure reports coverage of the reference map by the tested panel; class separation among the background domains is expected, since the descriptor set includes secondary-structure content and CATH class is defined on the same basis.}
\label{fig:expression_diversity}
\end{figure*}

The complete LAMBS production panel represented 612 published reference entries with 610 distinct Aether constructs, all with recorded expression means. The median expression concentration was 29.8 μM (Figure 7), and 417 constructs (68.4\%) were at or above 20 μM. The expression distribution includes proteins with undetermined binding labels. Of the 193 constructs omitted from the affinity–elution comparison, 107 had no accepted numeric affinities and 86 had insufficient affinity coverage for signal processing. Lower expression was common among these exclusions: 120 were below 20 μM. Concentration-related quality flags and the availability of determined affinities therefore both limited coverage of the correlation analysis.

\begin{figure}[tbp]
\centering
\includegraphics[width=\columnwidth]{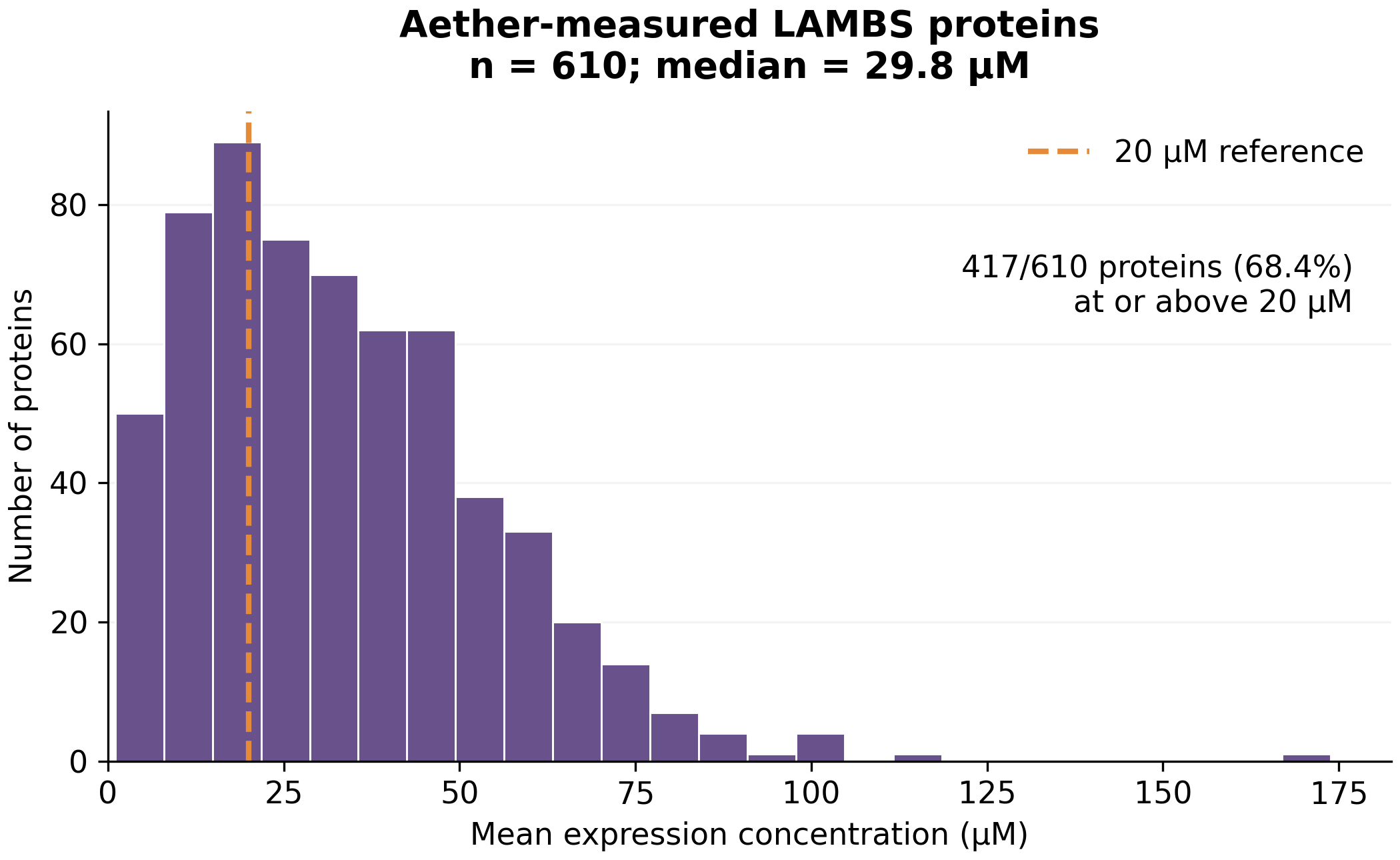}
\caption{Figure 7. Expression concentrations across the complete LAMBS production panel. Each of the 610 distinct Aether constructs contributes one registry expression mean, including constructs without determined binding affinities. The dashed orange line marks 20 μM; 417 constructs (68.4\%) meet or exceed this reference. The median concentration is 29.8 μM.}
\label{fig:lambs_expression_histogram}
\end{figure}

\section*{4. Discussion}
Screening-scale affinity measurements transfer to immobilized separation performance, and the transfer is carried by protein identity. Across 82 protein–metal-pair measurements, separation factors derived from primary $\mathrm{p}K_{\mathrm{d}}$ values tracked those measured on covalently immobilized protein under competition, with pairwise rank correlations of 0.878 to 0.923 and a pooled Pearson correlation of 0.836 against a permutation mean of 0.003, rising to 0.917 against 0.001 in the complete-panel test. The agreement spans a change from cell-free production to milligram-scale expression, from soluble protein to covalent immobilization, and from fitted dissociation constants to direct elemental quantification of eluted fractions. For candidate selection this is the result that matters: a protein ranked highly by the plate assay is ranked highly by the column.

The external comparison answers a different question and returns a different answer. Projections of the same primary affinities onto published LAMBS elution profiles reproduced those profiles at $r = 0.870$, but reassigning protein identity at random still produced $r = 0.840$. The lanthanide contraction imposes a shared ordering on every profile in the comparison, and that ordering accounts for the great majority of the apparent agreement. The protein-specific component is real and statistically detectable, but it is roughly 0.03 in correlation units. A reader given only the correlation coefficient would overestimate what the comparison demonstrates, as the authors initially did.

We draw a general point from this, stated as a caution rather than a criticism of any particular analysis. Where the measured entities form a chemically ordered family, correlations between normalized profiles are partly determined by the family's shared response ordering. This applies beyond the lanthanides: any analyte series with a dominant physical ordering, whether ionic radius, chain length, hydrophobicity or molecular weight, will produce agreement between profiles that describe different proteins. The magnitude of the effect is not predictable in advance; here it was 0.84 out of 0.87, and the only way to know was to measure it. A permutation of the identity labels is inexpensive, requires no additional experiment, and converts a correlation coefficient into a statement about how much of the agreement depends on the comparison being correctly matched. We would now ask about the endpoint first. A contrast at a fixed analyte pair cannot be inflated by the shared ordering, so it can be read as it stands. A normalized profile can be, by an amount nothing predicts, so it needs the null before its coefficient means anything.

The pairwise results make the same point more sharply. Dy/Pr and Pr/Nd rank correlations against LAMBS are modest in absolute terms, 0.331 and 0.610, yet both exceed every one of 1,000 scrambled values and therefore carry protein-specific information. Tb/Dy fails differently: its coefficient is also small, but it sits within the scrambled distribution ($p = 0.804$), so the comparison provides no evidence about that pair in the LAMBS format, in either direction. The pooled result is the third case, large and almost entirely shared. The absolute size of a correlation and the information it carries about individual proteins are distinct quantities, and in a chemically ordered series they can diverge in both directions.

Several limits bound these results. The external comparison is restricted to a single protein family, and the proteins advanced to secondary screening were selected on primary screening results rather than drawn at random, so the 35-protein cohort is not a random sample of the screening library. The immobilized measurements are benchtop column separations, not a production process; they change production scale, presentation and competition conditions relative to the primary assay, but they do not test flow regime, cycling, fouling or lifetime. Demonstrating transfer to those conditions requires their own measurements and, by the argument above, their own nulls.

The production platform showed broad coverage across the sampled soluble protein structures. Expression of non-LanM proteins spanning mainly α-helical, mainly β-sheet and mixed α/β classes extends its demonstrated scope beyond a single protein family. This broadens the structural range available for experimental screening and for the generation of functional training data.

\section*{5. Conclusions}
Structure-based confidence is a poor filter for protein function, which places the burden of design selection on measurement [2]. That burden is only discharged if the measurement remains informative under the conditions where the protein is used. Here we show that screening-scale affinity measurements rank proteins the same way an immobilized competitive separation does, with the correspondence resting on protein identity rather than on chemistry shared across the rare-earth series.

We also show that the usual evidence for such a claim can be misleading. Correlations between normalized profiles across a chemically ordered series reach 0.84 when protein identities are reassigned at random. The two comparisons reported here differ in endpoint, cohort and projection. The endpoint is what separates their results, and nothing predicts in advance how far a normalized profile will be inflated. Where an endpoint admits that inflation, a permutation of identity labels measures it at no experimental cost. For machine-learning-guided design, this is what determines whether a label is informative about the quantity a model is being trained to predict.

\end{document}